\documentclass[10pt,twocolumn,letterpaper]{article}

\usepackage{3dv}
\usepackage{times}
\usepackage{epsfig}
\usepackage{graphicx}
\usepackage{amsmath}
\usepackage{amssymb}

\usepackage{xcolor,colortbl}
\usepackage{comment}
\usepackage{multicol}
\usepackage{multirow}
\usepackage{caption}

\usepackage{epsfig}
\usepackage{float}
\usepackage{soul}
\usepackage{multicol}
\usepackage{blindtext}
\usepackage{diagbox}
\usepackage{makecell}
\usepackage{booktabs}
\usepackage{xspace}
\usepackage{balance}
\usepackage{color}
\usepackage{arydshln}
\usepackage{hyperref}

\usepackage{enumitem}

\usepackage[numbers,sort,compress]{natbib}
\usepackage{newfloat}
\DeclareFloatingEnvironment[
    fileext=los,
    listname={List of appendix figures},
    name=Figure,
    placement=tbhp,
    within=section,
]{sfig}

\makeatletter
\def\ps@myheadings{%
    \let\@oddfoot\@empty\let\@evenfoot\@empty
    \def\@evenhead{\thepage\hfil\slshape\leftmark}%
    \def\@oddhead{{\slshape\rightmark}\hfil\thepage}%
    \let\@mkboth\@gobbletwo
    \let\sectionmark\@gobble
    \let\subsectionmark\@gobble
    }
  \if@titlepage
  \renewcommand\maketitle{\begin{titlepage}%
  \let\footnotesize\small
  \let\footnoterule\relax
  \let \footnote \thanks
  \null\vfil
  \vskip 60\p@
  \begin{center}%
    {\LARGE \@title \par}%
    \vskip 3em%
    {\large
     \lineskip .75em%
      \begin{tabular}[t]{c}%
        \@author
      \end{tabular}\par}%
      \vskip 1.5em%
    {\large \@date \par}%
  \end{center}\par
  \@thanks
  \vfil\null
  \end{titlepage}%
  \setcounter{footnote}{0}%
}
\else
\renewcommand\maketitle{\par
  \begingroup
    \renewcommand\thefootnote{\@fnsymbol\c@footnote}%
    \def\@makefnmark{\rlap{\@textsuperscript{\normalfont\@thefnmark}}}%
    \long\def\@makefntext##1{\parindent 1em\noindent
            \hb@xt@1.8em{%
                \hss\@textsuperscript{\normalfont\@thefnmark}}##1}%
    \if@twocolumn
      \ifnum \col@number=\@ne
        \@maketitle
      \else
        \twocolumn[\@maketitle]%
      \fi
    \else
      \newpage
      \global\@topnum\z@   %
      \@maketitle
    \fi
    \thispagestyle{plain}\@thanks
  \endgroup
  \setcounter{footnote}{0}%
}
\makeatother

\threedvfinalcopy %

\def\threedvPaperID{*} %

\ifthreedvfinal\fi
\begin{document}

\newcommand{\rgb}{RGB\xspace}

\newcommand{\oneD}{{1D}\xspace}
\newcommand{\twoD}{{2D}\xspace}
\newcommand{\threeD}{\xspace{3D}\xspace}
\newcommand{\pytorch}{\mbox{PyTorch}\xspace}
\newcommand{\etend}{{end-to-end}\xspace}
\newcommand{\sota}{{state-of-the-art}\xspace}
\newcommand{\inthewild}{{in-the-wild}\xspace}
\newcommand{\mano}{\mbox{MANO}\xspace}
\newcommand{\groundtruth}{{ground-truth}\xspace}

\newcommand{\hod}{\mbox{HO-3D}\xspace}
\newcommand{\doh}{\mbox{100DOH}\xspace}
\newcommand{\dexycb}{\mbox{DexYCB}\xspace}
\newcommand{\freiH}{\mbox{FreiHAND}\xspace}
\newcommand{\hanco}{\mbox{HanCo}\xspace}
\newcommand{\peclr}{\mbox{PeCLR}\xspace}

\newcommand{\epe}{\mbox{EPE}\xspace}
\newcommand{\vtov}{\mbox{V2V}\xspace}
\newcommand{\PAvtov}{\mbox{PA-V2V}\xspace}
\newcommand{\RAvtov}{\mbox{RA-V2V}\xspace}
\newcommand{\PAepe}{\mbox{PA-EPE}\xspace}
\newcommand{\RAepe}{\mbox{RA-EPE}\xspace}
\newcommand{\STAepe}{\mbox{STA-EPE}\xspace}

\newcommand{\timecontrastive}{{time-contrastive}\xspace}
\newcommand{\methodname}{\mbox{TempCLR}\xspace}
\newcommand{\myparagraph}[1]{\vskip 0.6em\noindent\textbf{#1:}}

\newcommand{\tochange}[1]{\textcolor{red}{#1}}
\newcommand{\norm}[1]{\left\lVert#1\right\rVert}

\newcommand{\reffig}[1]{Fig.~\ref{#1}}
\newcommand{\refFig}[1]{Figure~\ref{#1}}
\newcommand{\reftab}[1]{Tab.~\ref{#1}}
\newcommand{\refTab}[1]{Table~\ref{#1}}
\newcommand{\refeq}[1]{Eq.~\ref{#1}}
\newcommand{\refEq}[1]{Equation~\ref{#1}}
\newcommand{\refsec}[1]{Sec.~\ref{#1}}
\newcommand{\refSec}[1]{Section~\ref{#1}}
\newcommand{\ccite}[1]{~\cite{#1}}
\newcommand{\rulesep}{\unskip\ \vrule\ }
\newcommand{\supmat}{\textcolor{black}{\textbf{SupMat}}\xspace}
\newcommand{\red}[1]{{\color{red}#1}}

\makeatletter
\def\blfootnote{\gdef\@thefnmark{}\@footnotetext}
\makeatother

\newcommand\scalemath[2]{\scalebox{#1}{\mbox{\ensuremath{\displaystyle #2}}}}

%%%%%%%%% TITLE
\title{\methodname: Reconstructing Hands via Time-Coherent Contrastive Learning}

\author{Andrea Ziani$^{1*}$ \quad Zicong Fan$^{1,2*}$ \quad Muhammed Kocabas$^{1,2}$ \quad Sammy Christen$^{1}$ \quad Otmar Hilliges$^{1}$\vspace{0.1cm} \\
 $^1$ETH Z{\"u}rich, Switzerland \quad
 $^2$Max Planck Institute for Intelligent Systems, T{\"u}bingen
}

\twocolumn[{%
\renewcommand\twocolumn[1][]{#1}%
\maketitle
\begin{center}
  \newcommand{\teaserwidth}{\textwidth}
  \vspace{-5mm}
  \centerline{\includegraphics[width=0.95\linewidth]{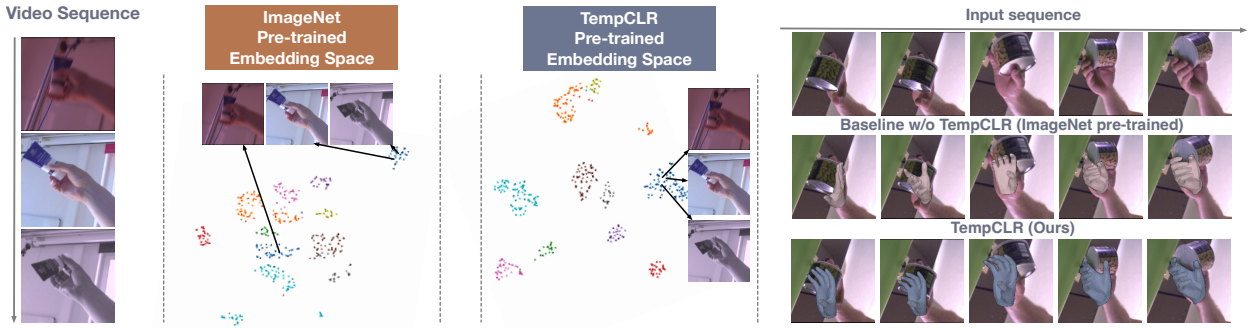}}
  \vspace{-8mm}
    \captionof{figure}{
    State-of-the-art hand reconstruction methods such as\ccite{choutas2020expose} (middle), fail to keep coherent hand representations through time. We exploit the underlying temporal constraint in unlabelled videos and train a model in a \timecontrastive manner. Our method (\methodname) keeps embeddings of the same sequence closer in the latent space and achieves better generalization on unseen videos, reconstructing more coherent hands through time.
    }
\label{fig:teaser}
\end{center}%
}]

\thispagestyle{empty}

\begin{abstract}
We introduce \methodname, a new time-coherent contrastive learning approach for  the structured regression task of \threeD hand reconstruction. 
Unlike previous \timecontrastive methods for hand pose estimation, our framework considers temporal consistency in its augmentation scheme, and accounts for the differences of hand poses along the temporal direction.
Our data-driven method leverages unlabelled videos and a standard CNN, without relying on synthetic data, pseudo-labels, or specialized architectures.
Our approach improves the performance of fully-supervised hand reconstruction methods by 15.9\% and 7.6\% in PA-V2V on the \hod and \freiH datasets respectively, thus establishing new \sota performance.
Finally, we demonstrate that our approach produces smoother hand reconstructions through time, and is more robust to heavy occlusions compared to the previous \sota which we show quantitatively and qualitatively.
Our code and models will be available at \href{https://eth-ait.github.io/tempclr}{https://eth-ait.github.io/tempclr}.
\end{abstract}

\blfootnote{*Equal contribution}

\section{Introduction}
Methods for hand pose and shape reconstruction have many applications in human-computer interaction, augmented reality, virtual reality, robotics, and motion generation\ccite{GRAB:2020,Taheri:CVPR:2022,christen2022dgrasp}.
Recent research demonstrates impressive results on the task of \emph{supervised} \threeD hand reconstruction from monocular \rgb images (\eg \ccite{romero2017embodiedhands, Zhang2019endtoend, hasson2019obman}).
However, generalizing to \inthewild settings, with fully unconstrained and uncontrollable environmental conditions, would require large amounts of training data captured under the same conditions.
As of today, accurate \threeD keypoint annotation of \inthewild data is an open research problem and, therefore, no large-scale \inthewild dataset with accurate 3D annotations exists.
For these reasons, techniques that leverage sparsely annotated data\ccite{Hasson2020photometric} or weakly labelled data\ccite{liu2021interaction, cao2021handobject, kulon2020youtube} have seen much interest. However, such methods rely on pseudo 2D or 3D annotations, which in turn require human effort for acquisition, or may introduce label noise that bounds model performance\ccite{cao2021handobject, kulon2020youtube}.
Therefore, a promising solution to avoid pseudo-labels entirely, is to make use of \emph{unlabelled} data, for example via contrastive learning\ccite{spurr2021self,zimmermann2021contrastive}.
In the context of sequence data, we observe that existing methods often struggle with heavy occlusions, for instance brought on by hand-object interaction. Consider the example from \reffig{fig:teaser}: while the hand pose throughout the grasp is quasi-static, the images change drastically from frame to frame, which causes existing methods to output incorrect hand poses.
In this paper, we explore how to learn better representations that capture human motion's inherent temporal consistency, improving the hand reconstruction stability through time. We do so by leveraging \textit{single-view} \emph{unlabelled} videos of hands grasping objects to improve 3D hand reconstruction in the most challenging setting of heavy occlusions. 

Unlike single images, videos contain temporal information that can help to predict coherent hand reconstructions through time by learning correlations between time-adjacent frames.
Combining this idea with the recent progress of contrastive representation learning methods\ccite{sermanet2018timecontrastive,chen2020simple, he2020momentum}, we introduce a \textit{time-coherent contrastive learning} pipeline, dubbed \methodname. Our approach consists of two stages, as shown in \refFig{fig:general_architecture}: a pre-training stage where we perform \textit{time-coherent contrastive learning} on unlabelled videos and a second stage, where the pre-trained encoder is fine-tuned on the \threeD hand reconstruction task using labelled data. In particular, \methodname contributes two key ideas: 1) a time-coherent augmentation method to impose strong spatial augmentations on each frame of a video while maintaining temporal integrity; and 2) a probabilistic sampling strategy that accounts for the differences in frames along the temporal dimension.
In contrast to a vanilla \timecontrastive learning approach\ccite{zimmermann2021contrastive}, which repels any non-neighboring frame in a sequence, our sampling strategy takes into consideration that temporally-closer frames often represent more  similar hand poses in range of motion.
Based on this insight, \methodname gives more attention to attracting temporally close frames and only repels temporally distant frames.
\refFig{fig:teaser} shows that our approach is able to produce smoother hand reconstructions along time, where a \sota approach\ccite{choutas2020expose} fails to do so.

We evaluate \methodname in different settings and on different datasets.
First, we demonstrate that our pre-training improves the performance over a fully-supervised baseline\ccite{choutas2020expose} by $15.9\%$ and $7.6\%$ in \threeD mesh error on the \hod and \freiH datasets (\cf \reftab{tab:sota_ho3d} and \reftab{tab:sota_freihand}).
Next, we show that our \textit{single-view} time-contrastive method improves over a vanilla \timecontrastive approach\ccite{zimmermann2021contrastive} on \freiH. 
Through cross-dataset evaluation and \inthewild qualitative results, we show improvements in generalization capabilities.
Finally, we demonstrate that our method yields smoother hand reconstructions along the temporal dimension compared to other SotA approaches.

Our contributions can be summarized as follows: 
\vspace{-1mm}
\begin{enumerate}[noitemsep]
   \item A novel \textit{single-view} time-contrastive learning approach for 3D hand reconstruction. Our method leverages time-coherent augmentations and a probabilistic sampling strategy to capture long-range dependencies.
   \item We experimentally show that by leveraging \inthewild unlabelled monocular videos, \methodname outperforms existing methods across different metrics.
    \item We provide empirical evidence that our method leads to smoother hand poses estimated over time.
\end{enumerate}

\section{Related Work}

\myparagraph{Fully-supervised 3D hand reconstruction} 
Reconstructing hands in \threeD from images has received increased attention in recent years\ccite{Supancic2015surveyHands,Yuan2018surveyHands}.
Existing methods\ccite{grady2021opt,iqbal2018heatmap,mueller2018ganerated,fan2021digit,simon2017keypoint,zimmermann2017learning,Boukhayma2019,Zhang2019endtoend,moon2020interhand, hasson2019obman,spurr2020eccv,tekin2019egocentric,he2020epipolar,ge2019hand} often leverage full supervision from in-the-lab datasets.
For instance, Zimmermann \etal \cite{zimmermann2017learning} propose the first convolutional network to detect \twoD hand joints and lift them into the \threeD space with an articulation prior.
Iqbal \etal \cite{iqbal2018heatmap} introduce a 2.5D representation for \threeD hand pose estimation. 
Boukhayma \etal\ccite{Boukhayma2019} and Choutas \etal\ccite{choutas2020expose} estimate MANO\ccite{romero2017embodiedhands} and SMPL-X\ccite{pavlakos2019expressive} parameters using a weak perspective camera model.
Lin \etal\ccite{lin2021metro} introduce a transformer architecture to estimate vertices of the MANO mesh.
In contrast to these approaches, we focus on leveraging additional supervision from unlabelled videos to improve \threeD hand reconstruction.

\myparagraph{Reconstructing hands from limited supervision}
Recently, several datasets for \threeD hand pose and shape estimation have been introduced \cite{hampali2020honotate, chao2021dexycb, garcia2018firstperson, zimmermann2019freihand, moon2020interhand,jin2020whole}. 
However, capturing \threeD hand annotation is difficult to scale:
1) Magnetic trackers\ccite{garcia2018firstperson} provide \threeD annotation for hands and objects but they are intrusive and introduce noise in RGB images.
2) Multi-view setups\ccite{hampali2020honotate, chao2021dexycb, moon2020interhand,brahmbhatt2020contactpose} are marker-less, but the labels are obtained by either manual \twoD annotation with triangulation\ccite{moon2020interhand,chao2021dexycb} or from noisy multi-kinect systems\ccite{hampali2020honotate}; the quantity of \threeD labelled data is still limited, and the background is not diverse.
3) Synthetic data provides perfect ground-truth but lacks photorealism\ccite{hasson2019obman,mueller2018ganerated}. 

To allow methods to generalize to unconstrained settings, recently, there has been attention on reducing the reliance on \threeD annotation\ccite{baek2019pushing, Zhang2019endtoend, cao2021handobject, Boukhayma2019,Hasson2020photometric,liu2021interaction,spurr2021self,zimmermann2021contrastive,spurr2018cross,spurr2021adversarial}.
For example, Hasson \etal \cite{Hasson2020photometric} leverage sparsely annotated data by introducing a photometric loss formulation to learn from partially labelled sequences. 
Liu \etal \cite{liu2021interaction} propose a specialized transformer-based architecture used to collect pseudo labels from \inthewild videos. These pseudo labels are then used to train the same architecture. 
Zimmermann \etal \cite{zimmermann2021contrastive} explore the benefits of multi-view and single-view \timecontrastive learning applied on the hand reconstruction task.
Spurr \etal \cite{spurr2021self} introduce an equivariant contrastive objective formulation where geometric transformations applied on the image are reversed in the latent space.
In this paper, we introduce a self-supervised approach to leverage supervision on unlabelled monocular videos in the wild. 

The most relevant methods to us are\ccite{Hasson2020photometric, liu2021interaction,spurr2021self,zimmermann2021contrastive}, which leverage unlabelled or partially labelled data. 
Compared to\ccite{Hasson2020photometric,liu2021interaction}, our method requires neither human intervention for tuning pseudo-labels\ccite{liu2021interaction}, nor sparsely annotated videos\ccite{Hasson2020photometric}.
Similarly to ours, the methods in\ccite{spurr2021self,zimmermann2021contrastive} use a contrastive formulation.
However,\ccite{spurr2021self} relies on unlabelled in-the-wild still images while we rely on unlabelled in-the-wild videos. 
In addition,\ccite{zimmermann2021contrastive} leverage a multi-view time-contrastive formulation while our approach is based on monocular videos.
Furthermore, to go beyond\ccite{zimmermann2021contrastive}, we introduce a simple-yet-effective time-coherent augmentation method and sampling strategy that reflects the differences in frames along time. 
Experiments show that this novel combination is crucial for time-contrastive learning.

\section{Method}\label{sec:method}
\begin{figure*}[t]
\centering
\includegraphics[width=\linewidth]{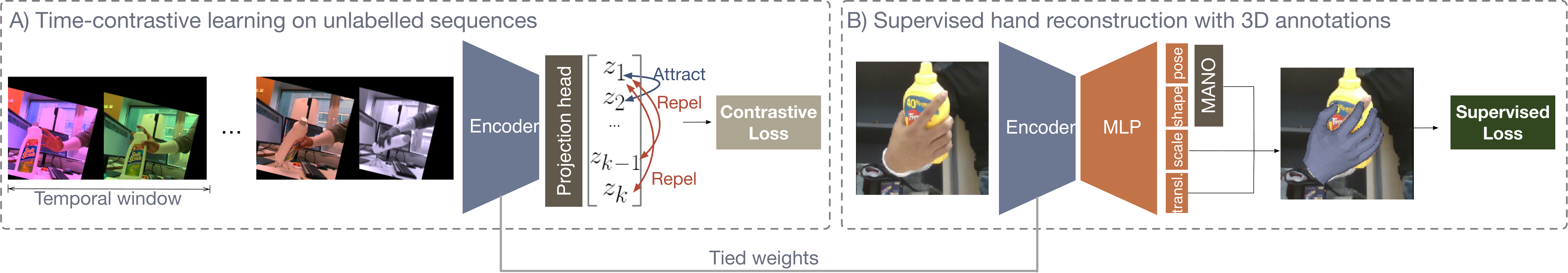}
\caption{\textbf{Overview of \methodname}: A) An encoder is trained with a \timecontrastive learning approach on unlabelled videos of hands grasping objects. B) The pre-trained encoder is fine-tuned using labelled data.}
\label{fig:general_architecture}
\end{figure*}

\refFig{fig:general_architecture} shows a schematic of our method, \methodname,
which consists of two stages: a pre-training stage, and a fine-tuning stage. In the pre-training stage, we leverage a time-contrastive objective to train the image encoder on unlabelled videos.
This stage is to obtain additional supervision for the encoder from diverse \inthewild videos of hand in motion.
In the second stage, we train the whole hand reconstruction architecture through supervised fine-tuning.
In \refSec{sec:method:contrastive_learning}, we describe our time-contrastive pre-training,
motivating the importance of our data augmentation and probabilistic sampling technique. Then, in \refsec{sec:method:hand_reconstruction} we present our hand reconstruction model.

\subsection{Time-contrastive Learning}\label{sec:method:contrastive_learning}

We build our self-supervised time-contrastive learning framework as illustrated in \reffig{fig:general_architecture}A.
The core of our framework is an NT-Xent loss \cite{chen2020simple} applied on features extracted from augmented frames of a sequence (the augmentation module is described below).
We denote a video as $X = \{x_{1}, x_{2}, ..., x_{n}\}$, where $x_t$ is the \textit{t-th} frame of the sequence. Around a reference frame $x_i$, we define the temporal window $T_i = \{x_{i-k}, .., x_{i-1}, x_{i+1}, .., x_{i + k}\}$ with size $2k$. Frames inside this temporal range correspond to the candidate positive pairs of frame $x_i$, while all the other frames of the same video correspond to candidate negative pairs. We use $z_i$ to denote the encoded representation of $x_i$.

Suppose that we sample $M$ frames per mini-batch, possibly from different videos; for each frame $x_i$ we sample $P_i \subseteq T_i$ (positive pairs), and $N_i \subseteq X \setminus T_i$ (negative pairs). $|P_i|$ and $|N_i|$ are fixed.
The NT-Xent loss is defined as:
\begin{gather}
  L = \frac{1}{M} \sum_{i=1}^{M}L_i,\\
L_i = - \sum_{x_j \in P_i} \log \frac{\exp{(sim(z_i, z_j)/ \tau)}}{\sum_{x_k \in N_i} \exp{(sim(z_i, z_k)/ \tau)}} .
\end{gather}
Here, $\tau > 0$ is a temperature parameter and $sim(u, v) = u^{T} v / \norm{u} \norm{v}$ is the cosine similarity between $z_i$ and $z_j$. 
Hence, the loss encourages embeddings of similar, neighboring frames (positive pairs) to be mutually attracted while those of dissimilar frames in the same sequence (negative pairs) are kept far apart.

\begin{figure}[]
\centering
\includegraphics[width=0.85\linewidth]{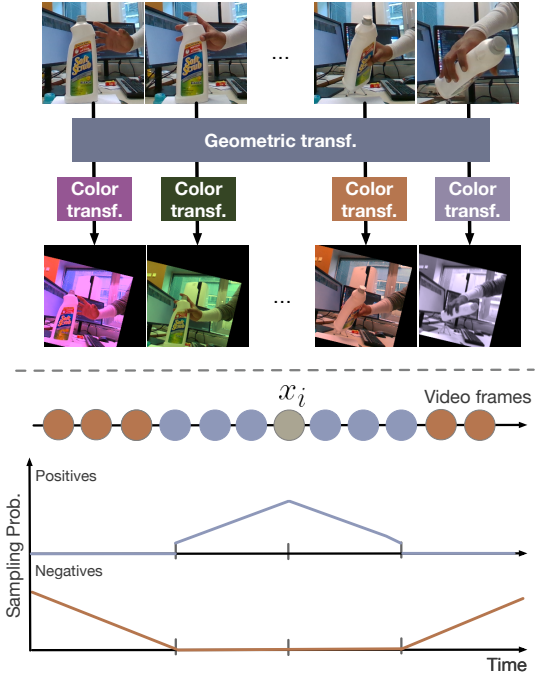}
\caption{\textbf{Overview of the time-coherent augmentations (top) and the probabilistic sampling step (bottom)}. }
\vspace{-6mm}
\label{fig:mini_batch_construction}
\end{figure}

\myparagraph{Time-coherent geometric transformations}
Data augmentations are extensively used in contrastive training for computer vision tasks\ccite{chen2020simple, he2020momentum, qian2021spatiotemporal}.
Although a common optimal augmentation procedure does not exist, in a temporal setting a natural approach is to employ existing augmentation methods to the frames of the video one by one. 
Image augmentation methods often include geometric transformations such as random cropping, rotation, translation. 
In sequences, however, such transformations could break the inherent motion cues between consecutive frames, negatively affecting representation learning along the temporal dimension.
Inspired by Qian \etal\ccite{qian2021spatiotemporal}, we apply consistent augmentations through time by applying the same random geometric transformations (\ie rotation, scale, and translation) across frames of the same sequence, while applying independent appearance transformation for each frame (see \refFig{fig:mini_batch_construction}). In this way, the encoder better captures temporal features in the pre-training stage. 

\myparagraph{Probabilistic pair sampling} Existing method on \timecontrastive learning for hand pose estimation \cite{zimmermann2021contrastive} defines two immediate neighbouring frames as positive pairs and any couple of non-neighbouring frames as negatives pairs. In the grasping scenario, however, the hand pose has a limited range of movement caused by the interaction between the hand and the object. This means that several consecutive frames could represent similar hand poses and a trivial pair selection may not be beneficial.
To address this problem, our key insight is that two images from the same video represent more diverse hand poses when their temporal distance is large. 
To this end, we use a sampling strategy to account for the temporal changes (see \reffig{fig:mini_batch_construction}).
In particular, given a frame $x_i$ sampled uniformly at random from a sequence, we first define a temporal window $T_i$, as described in the previous section. 
Then, from the temporal window, we sample $P_i$ positive pairs with a probability distribution that monotonically decreases with the distance from $x_i$. Likewise, we sample $N_i$ negative pairs, lying outside the temporal window, with a probability directly proportional to the distance from $x_i$. 
Following our sampling strategy, the contrastive training will focus more on attracting temporally closer frames and repelling temporally more distant frames, while reducing the attention that is given to the grey zone of frames representing hand poses with uncertain similarity to $x_i$.

To summarize our pre-training approach, first, each frame of a sequence is augmented by the same geometric transformation. Then, each frame is augmented independently via random (potentially different) appearance augmentations. After that, the sampling strategy chooses the positive and negative frames. See \supmat for more details.

\subsection{Hand Reconstruction}\label{sec:method:hand_reconstruction}
\refFig{fig:general_architecture}B shows our hand reconstruction network.
Following\ccite{choutas2020expose,Hasson2020photometric,Boukhayma2019}, we use an encoder-decoder formulation.
In particular, our method consists of our pre-trained encoder to obtain an image feature vector and a hand decoder to predict the MANO pose and shape parameters and the weak perspective camera parameters (scale and translation).
Formally, given an image, the network predicts the MANO  pose vector $\theta = [\theta^{\text{wrist}}; \theta^{\text{fingers}}] \in \mathbb{R}^{16 \times \text{D}}$, shape parameters $\beta \in \mathbb{R}^{10}$ and the weak perspective camera parameters ($t, s$). The MANO parameters are fed into the MANO differentiable layer to retrieve the 3D hand mesh.
The weak perspective camera model aligns the mesh with the image.
Following\ccite{choutas2020expose}, we use the rotation representation from Zhou \etal \cite{zhou2019continuity} for our MANO pose parameters ($D=6$).

We train our model using 2D re-projection loss, 3D joint errors, and pose and shape parameter loss $L = \lambda_{2\text{D}} L_{2\text{D}} + \lambda_{3\text{D}} L_{3\text{D}} + \lambda_{\Theta} L_{\Theta}$,
  where $L_{2\text{D}} = ||J^{2\text{D}} - \hat{J}^{2\text{D}}||_1$, $L_{3\text{D}} = ||J^{3\text{D}} - \hat{J}^{3\text{D}}||_1$, and $L_{\Theta} = ||\{\theta, \beta\} - \{\hat{\theta}, \hat{\beta}\}||^{2}_2$.
All variables with a hat denote predictions and $J^{2\text{D}} \in \mathbb{R}^{21 \times 2}$ and $J^{3\text{D}} \in \mathbb{R}^{21 \times 3}$ represent the 21 keypoints in 2D and 3D.

\section{Experiments}\label{sec:experiments}
In \refSec{sec:experiments:details}, we first introduce experiment details such as the datasets, the evaluation metrics, and the implementation details. 
In \refsec{sec:experiments:sota_comparison}, we compare our method to \sota approaches on both hand-grasping-objects and hand-only settings.
In \refsec{sec:experiments:method_analysis}, we ablate \methodname and provide qualitative results. Also, we show the effectiveness of \methodname when \threeD annotations are scarce.
Finally, in \refsec{sec:experiments:cross_eval}, we perform cross-dataset evaluation to demonstrate generalization under domain shifts.

\subsection{Datasets, Metrics, and Implementation Details}
\label{sec:experiments:details}
\myparagraph{\hod \ccite{hampali2020honotate}} The dataset provides \threeD hand-object annotations during interaction for markerless \rgb images. The ground-truth annotations are obtained by fitting a hand model to multi-view RGB-D evidence. We present results on \hod v2. The evaluation is performed online; hence we do not have access to the ground truth of the test set.
\myparagraph{\freiH \cite{zimmermann2019freihand}, \hanco \ccite{zimmermann2021contrastive}} 
\freiH (FH) consists of 130k training and 4k evaluation samples captured with a green screen background in the training set, as well as real backgrounds in the test set. Both \threeD and \twoD annotations are provided.
\hanco does not contain \threeD annotations. It only contains short video clips recorded with a calibrated and time-synchronized multi-view camera capture setup.
In total, there are 107k time-steps recorded by eight cameras, which results in 860k \rgb images.
As these datasets are composed of both hand-only and hand-grasping-object sequences, we used \hanco in the time-contrastive pre-training and FH in supervised fine-tuning. 
\myparagraph{100 Days Of Hands \cite{shan2020internet}} This is a large-scale and in-the-wild dataset of hand-object interaction footage. The dataset does not provide any hand annotation besides the bounding boxes of the hands in the scene. 
In some of our experiments, we used a subset of 10 videos collected from this dataset (86k images) exclusively as additional unlabelled frames for time-contrastive pre-training. We show that this pre-training improves hand reconstruction.

\myparagraph{Evaluation metrics}
We report the End-Point-Error (\epe) and the Vertex-to-Vertex End-Point-Error (\vtov). The former denotes the average L2 distance between the ground-truth and predicted keypoints, while the latter denotes the average L2 distance between the ground-truth and mesh vertices. 
We prefix the metrics with PA, RA and STA to denote procrustes alignment, root alignment, and scale-and-translation alignment. 
We include the F-scores defined as the harmonic mean between recall and precision between two meshes given a distance threshold.
Following\ccite{kocabas2020vibe}, to measure the temporal stability of the reconstruction, we compute an acceleration error by measuring the difference in acceleration between the 3D GT and the predictions. 

\myparagraph{Implementation details}
For the pre-training we use ResNet\ccite{he2016resnet} as a backbone, which takes monocular \rgb images of size $224 \times 224$ as input. We employ Adam \cite{kingma2015adam} as the optimizer with a batch size of $2048$ and a learning rate of $4.5e^{-3}$ for $50$ epochs. 
The fine-tuning is performed until convergence based on the performance on the validation set.
During fine-tuning, we use \rgb images of size $224 \times 224$ as input. As optimizer, we use Adam with a learning rate of $5e^{-4}$ and a batch size of $128$. 
Further details can be found in \supmat.
Following\ccite{sermanet2018timecontrastive}, we choose the window size to be approximately half of the frame rate for each dataset (15 for HO-3D and 100DOH, 5 for HanCo).

\begin{table}[t]
\centering
\resizebox{\columnwidth}{!}{
\begin{tabular}{lccccc} 
 \hline
 \multirow{2}{*}{Method} & PA-V2V  & PA-EPE & F@5 & F@15\\
 & ($mm$) $\downarrow$ & ($mm$) $\downarrow$ & $mm$ $\uparrow$ & $mm$ $\uparrow$\\
 \hline\hline
 Baseline\ccite{choutas2020expose} & 12.6 & 12.7 &  0.389 & 0.905 \\
 \rowcolor{gray!15}Hasson \etal\ccite{Hasson2020photometric} & 11.4 & 11.4 & 0.428 & 0.932 \\
 Hasson \etal\ccite{hasson2019obman} & 11.2 & 11.1 & 0.464 & \textbf{0.939} \\
 \rowcolor{gray!15} \peclr\ccite{spurr2021self} & 10.8 & 10.8 & 0.47 & 0.936\\
   \textbf{TempCLR (ours)} & \textbf{10.6} & \textbf{10.6} & \textbf{0.481} & 0.937\\  \Xhline{2\arrayrulewidth}
 \rowcolor{gray!15}PeCLR$^\dagger$\ccite{spurr2021self} & 11.0 & 11.0 & 0.46 & 0.934\\
  \textbf{TempCLR$^\dagger$ (ours)}& \textbf{10.0} & \textbf{10.1} &  \textbf{0.505} & \textbf{0.947}\\
  \hdashline
 Liu \etal\ccite{liu2021interaction} & 9.5 & 9.9 & 0.526 & 0.955\\
 
 \hline
\end{tabular}
}
\caption{\textbf{Comparison with SotA on \hod\ccite{hampali2020honotate}}.
\methodname outperforms the baselines on all reported metrics. 
The employment of additional \inthewild data for contrastive pre-training, denoted by $\dagger$, further improves the model's accuracy. 
Results below the dashed line employ weak supervision.}
\label{tab:sota_ho3d}
\end{table}
\subsection{Comparison with the State-of-the-Art}\label{sec:experiments:sota_comparison}
Here we compare \methodname with fully-supervised and self-supervised \sota approaches on \hod and FH.
\refFig{fig:qualitative_res} shows qualitative results.

\myparagraph{Comparison on \hod}
\refTab{tab:sota_ho3d} compares \methodname with the fully-supervised and self-supervised \sota on \hod. 
First, we pre-train a ResNet18 encoder on unlabelled \hod images. Then, we fine-tune the hand reconstruction network with full supervision as described in \refsec{sec:method:hand_reconstruction}. 
To show that our self-supervised method can leverage \inthewild unlabelled data, we repeat the experiment but include additional unlabelled frames from \doh, along with the original unlabelled frames in \hod, during the contrastive training phase. 

Top rows of the table show that \methodname, without employing any \inthewild data, improves over our fully-supervised baseline\ccite{choutas2020expose} (see Baseline on the table) by $15.9\%$ in \PAvtov and \PAepe. 
Furthermore, using additional \inthewild data for \timecontrastive pre-training (denoted by $\dagger$ in \reftab{tab:sota_ho3d}), \methodname improves further and establishes the new \sota for self-supervised training.
Notably, \methodname is on par with\ccite{liu2021interaction}, a weakly-supervised method that uses pseudo-labels. The labels involve manual intervention to generate. \methodname is self-supervised, so it does not require intervention to train on unlabelled videos.

With additional \inthewild data, \peclr pre-training does not further improve.
This is consistent to the observation in Fig. 6 of the PeCLR paper --
although PeCLR improves hand poses by leveraging additional in-the-wild data (FH+YT3D) compared to fully-supervised training (FH), the improvement is significant in \textit{low data regime}.
With more annotation, training with additional in-the-wild data does not lower the error.
In contrast, our method consistently improves over the baseline in both low data and high data regime (see \reftab{tab:sota_ho3d} and \reffig{fig:contrastive_scarce_data}).

\begin{table}[t]
\centering
\resizebox{\columnwidth}{!}{
\begin{tabular}{lccccc} 
 \hline
 \multirow{2}{*}{Method} & PA-V2V  & RA-V2V & F@5 & F@15\\
 & ($mm$) $\downarrow$ & ($mm$) $\downarrow$ & ($mm$) $\uparrow$ & ($mm$) $\uparrow$\\
 \hline\hline  
    Hasson \etal\ccite{hasson2019obman}  & 13.2 & - & 0.436 & 0.908 \\
   \rowcolor{gray!15}
   Baseline-18\ccite{choutas2020expose} &  11.8 & 35.96 & 0.484 & 0.918 \\
 \textbf{TempCLR-18 (ours)} &
 \textbf{10.9} & \textbf{25.05} & \textbf{0.513} & \textbf{0.930} \\
 \Xhline{2\arrayrulewidth}
 \rowcolor{gray!15}
   Baseline-50\ccite{choutas2020expose} & 10.8 & 31.15 & 0.518 & 0.934 \\
  MANO CNN\ccite{zimmermann2019freihand} &  10.7 & - & 0.529 & 0.935 \\
  \rowcolor{gray!15}
  HanCo Augm.\ccite{zimmermann2021contrastive} & 10.9 & - & 0.521 & 0.934 \\
    HanCo Temporal\ccite{zimmermann2021contrastive} & 10.4 & - & 0.538 & 0.939 \\
    \rowcolor{gray!15}PeCLR-50\ccite{spurr2021self} & 10.6 & 26.73 & 0.533 & 0.940 \\
  \textbf{TempCLR-50 (ours)} &  \textbf{10.2} & \textbf{21.68} & \textbf{0.541} & \textbf{0.941} \\
  \hdashline
  HanCo Multi-view\ccite{zimmermann2021contrastive} & 10.2 & - & 0.548 & 0.943 \\
 \hline
\end{tabular}
}
\vspace{-0.1cm}
\caption{\textbf{Comparison with SotA on FH\ccite{zimmermann2019freihand}}.
The top-bottom split (solid line) separates results using ResNet18 and ResNet50. 
The dashed line separates a multi-view temporal approach that is not directly comparable.
}
\label{tab:sota_freihand}
\end{table}
\myparagraph{Comparison on FH}
Here we use the \hanco dataset alone to perform our contrastive pre-training on ResNet18 and ResNet50 encoders. To show the efficacy of \methodname, we compare the results produced by our pipeline against fully-supervised methods and \sota contrastive approaches\ccite{zimmermann2021contrastive,spurr2021self}.
Before diving into results, we highlight that we report the \RAvtov scores for the fully-supervised baseline (ExPose\ccite{choutas2020expose}) and for our \timecontrastive approach only. 
This is because the FH test set was previously hidden and hosted as competition online, where this metric was not computed. 
Moreover, we do not have access to the pre-trained models to reproduce the missing results.

\refTab{tab:sota_freihand} shows that \methodname improves over the ResNet18 fully-supervised baseline by $30.4\%$ in \RAvtov and by $7.6\%$ in \PAvtov, indicating a significant improvement in global orientation and scale.
Similarly, with a ResNet50 backbone, \methodname improves over the baseline by $30.4\%$ in \RAvtov and by $5.5\%$ in \PAvtov.
Finally, we establish \sota performance by improving over the single-view self-supervised approach\ccite{zimmermann2021contrastive}. Note that the \RAvtov metric is not available for \ccite{zimmermann2021contrastive}.
Our single-view \timecontrastive approach is on par with the multi-view \timecontrastive approach proposed by Zimmermann \etal\ccite{zimmermann2021contrastive}.
We emphasize that monocular videos are more abundant on the Internet and often have very diverse environments in comparison to controlled multi-view setup.

\subsection{Ablation Study}\label{sec:experiments:method_analysis}
Here we ablate our method, and support it with quantitative and qualitative results. First, we analyse the embedding space learned through \methodname pre-training and compare it to an ImageNet pre-trained encoder.
We ablate the importance of time-coherent augmentation and probabilistic sampling for time-contrastive learning, and we provide evidence that time-coherent contrastive learning leads to more stable hand reconstructions through time.
Next, we compare different probabilistic sampling strategies. 
Lastly, we evaluate the efficacy of \methodname when ground-truth data for fine-tuning is scarce.

\begin{figure*}[t]
\centering
\includegraphics[width=0.9\linewidth]{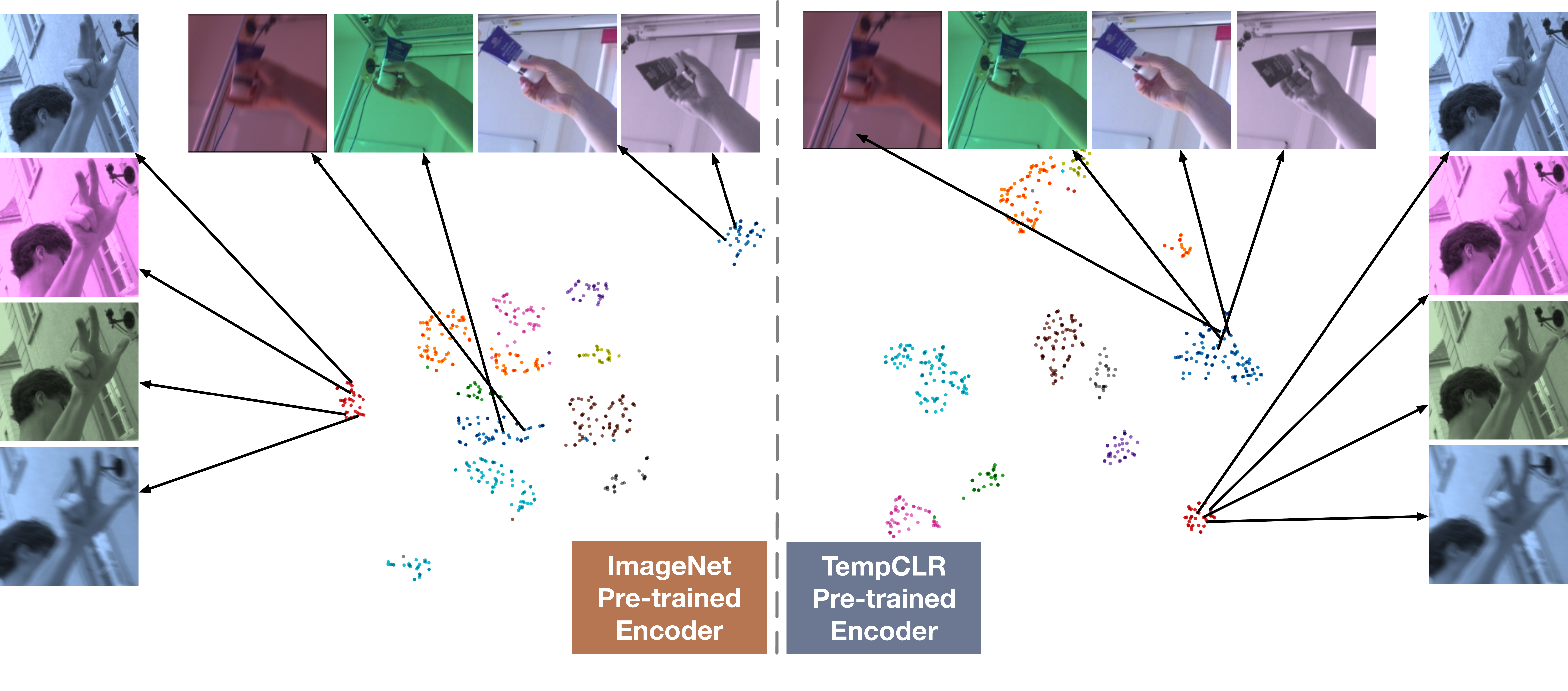}
\vspace{-4mm}
\caption{\textbf{Comparison of the 2D \textit{t-SNE} embeddings} produced by an encoder pre-trained on ImageNet and by our \timecontrastive pre-trained encoder. On the right hand side, we see that hand poses close along the temporal dimension are located in proximity to each other. Contrary, on the left hand side hand poses close in time are more distant in the embedding space.}
\label{fig:tsne}
\end{figure*}

\myparagraph{Latent space representation}
\refFig{fig:tsne} shows a t-SNE plot of two embedding spaces comparing the ImageNet pre-trained backbone and our backbone with a \methodname pre-training.
In particular, ten different sequences from the \hanco dataset have been randomly sampled and augmented. For each image of these sequences, we extract their feature vector and perform a t-SNE clustering. 
We see that \methodname leads to better cluster separation and, within the same cluster, similar hand poses are closer in the embedding space.
This confirms that our method yields the desired latent spaces we described in \refsec{sec:method}.

\begin{table}[t]
\centering
\resizebox{\columnwidth}{!}{
\begin{tabular}{lcccc} 
 \hline
 \multirow{2}{*}{Method} & Accel.  & \RAepe & \PAepe\\
 & ($mm/s^2$) $\downarrow$ & ($mm$) $\downarrow$ & ($mm$) $\downarrow$\\
 \hline
 Baseline\ccite{choutas2020expose} & 54.11 & 75.42 & 15.30\\
  \rowcolor{gray!15}  TempCLR w/o consist. augm. & 45.87 & 61.28 & 14.51\\
 TempCLR w/o prob. sampling & 47.56 & 52.31 & 13.80\\
  \rowcolor{gray!15}  \textbf{TempCLR} & \textbf{45.37} & \textbf{51.72} & \textbf{13.69}\\
 \hline
\end{tabular}
}
\vspace{-0.1cm}
\caption{\textbf{Temporal stability evaluation}.
Our augmentation strategy improves the hand pose estimation performance, while the sampling strategy helps in temporal stability. The combination of the two leads to the best results. 
}
\label{tab:augmentations_ablation}
\vspace{-4mm}
\end{table}
\begin{table}[t]
\centering
\begin{tabular}{lcccc} 
 \hline
 Augmentation & RA-EPE ($mm$) $\downarrow$ & PA-EPE ($mm$) $\downarrow$\\  
 \hline
 Rotation & 141.34 & 18.02 \\
 \rowcolor{gray!15} Translation & 99.04 & 15.93 \\
 Scale & 98.21 & 15.50\\
 \rowcolor{gray!15} Channel Noise & 96.76 & 15.70 \\
 Color Drop & 98.19 & 15.60\\
 \rowcolor{gray!15} Color Jitter & 97.36 & 15.76 \\
 Sobel Filter & 97.35 & 15.71 \\
 \hline
\end{tabular}
\caption{\textbf{Effects of different data augmentations in pre-training.} We pre-train on HanCo, fine-tune on FH, evaluate on FH test set. 
}
\label{tab:augmentation_ablation}
\end{table}
\begin{table}[t]
\centering
\begin{tabular}{lcccc} 
 \hline
 Method & RA-EPE ($mm$) $\downarrow$ & PA-EPE ($mm$) $\downarrow$\\  
 \hline
 Baseline\ccite{choutas2020expose} & 35.96 & 11.8\\
 \rowcolor{gray!15} \methodname-Lin & \textbf{25.05} & \textbf{10.9}\\
 \methodname-Exp & 28.91 & 11.4\\
 \rowcolor{gray!15} \methodname-Tanh & 28.85 & 11.1\\
  
 \hline
\end{tabular}
\caption{\textbf{Effects of different sampling strategies in pre-training.}
We pre-train on HandCo, fine-tune on FH, and evaluate on FH test set.
}
\vspace{-4mm}
\label{tab:sampling_ablation}
\end{table}

\myparagraph{Effects of time-coherent augmentation and probabilistic sampling}
We compare the fully-supervised baseline\ccite{choutas2020expose} trained on FH, and our method pre-trained on \hanco and fine-tuned on FH. In addition, we investigate the influence on the final performance of each of our contributions by removing our time-coherent geometric augmentation and the probabilistic sampling strategy (see \refsec{sec:method}).
Since \freiH is not a temporal dataset and the \hod test set is hidden, we evaluate on the \hod training split.
\refTab{tab:augmentations_ablation} shows that the greatest improvement in hand pose estimation (\RAepe and \PAepe) comes from the augmentation strategy, while the probabilistic sampling strategy contributes more to the temporal stability (see the acceleration metric).
These results confirm our insight that when performing \timecontrastive learning for images with hands in motion, it is crucial to sample distant frames to ensure the feasibility of the pre-training task.
The acceleration metric demonstrates that our pre-training leads to more stable results even using a single-frame model.
Moreover, the time-coherent geometric augmentation and the sampling strategy complement each other and the combination of the two leads to the best overall improvement.
See \supmat for additional qualitative results and failure cases.

\myparagraph{Different augmentation strategies}
\label{supp:augmentation_abl}
\refTab{tab:augmentation_ablation} shows the impact of different augmentations in the pre-training stage.
In particular, we pre-train on HanCo\ccite{zimmermann2021contrastive}, and fine-tune on FreiHAND\ccite{zimmermann2019freihand} with a ResNet18\ccite{he2016resnet} backbone. 
Similar to\ccite {spurr2021self}, the appearance transformations are more beneficial than geometric transformations. 
This motivates our choice to keep independent appearance transformations for each frame of a sequence while preserving the motion of the video with coherent geometric transformations in time. 

\begin{figure}[t]
\centering
\centering
\begin{tabular}{@{}c@{}}
     \includegraphics[width=0.9\linewidth]{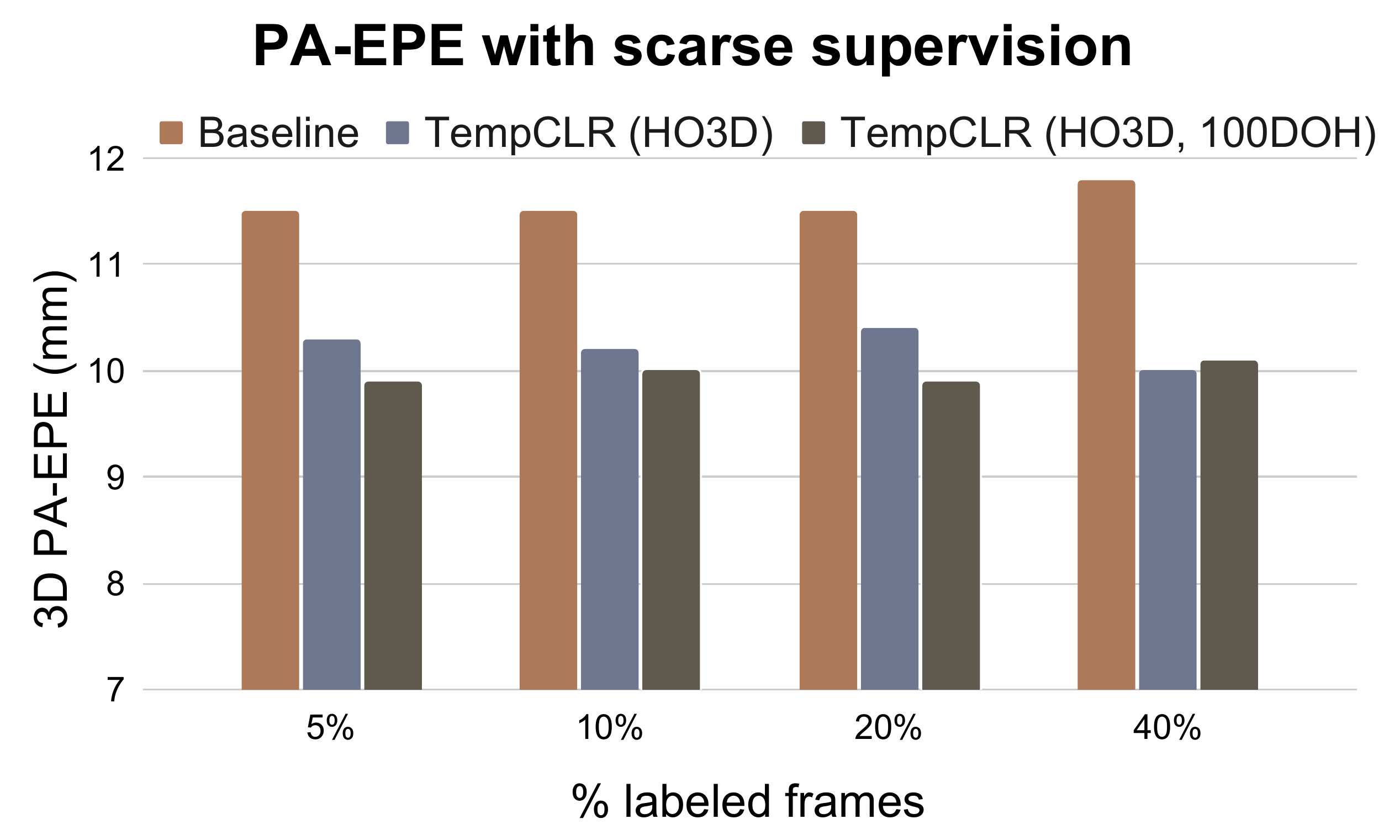}
\end{tabular}
\begin{tabular}{@{}c@{}}
     \includegraphics[width=0.9\linewidth]{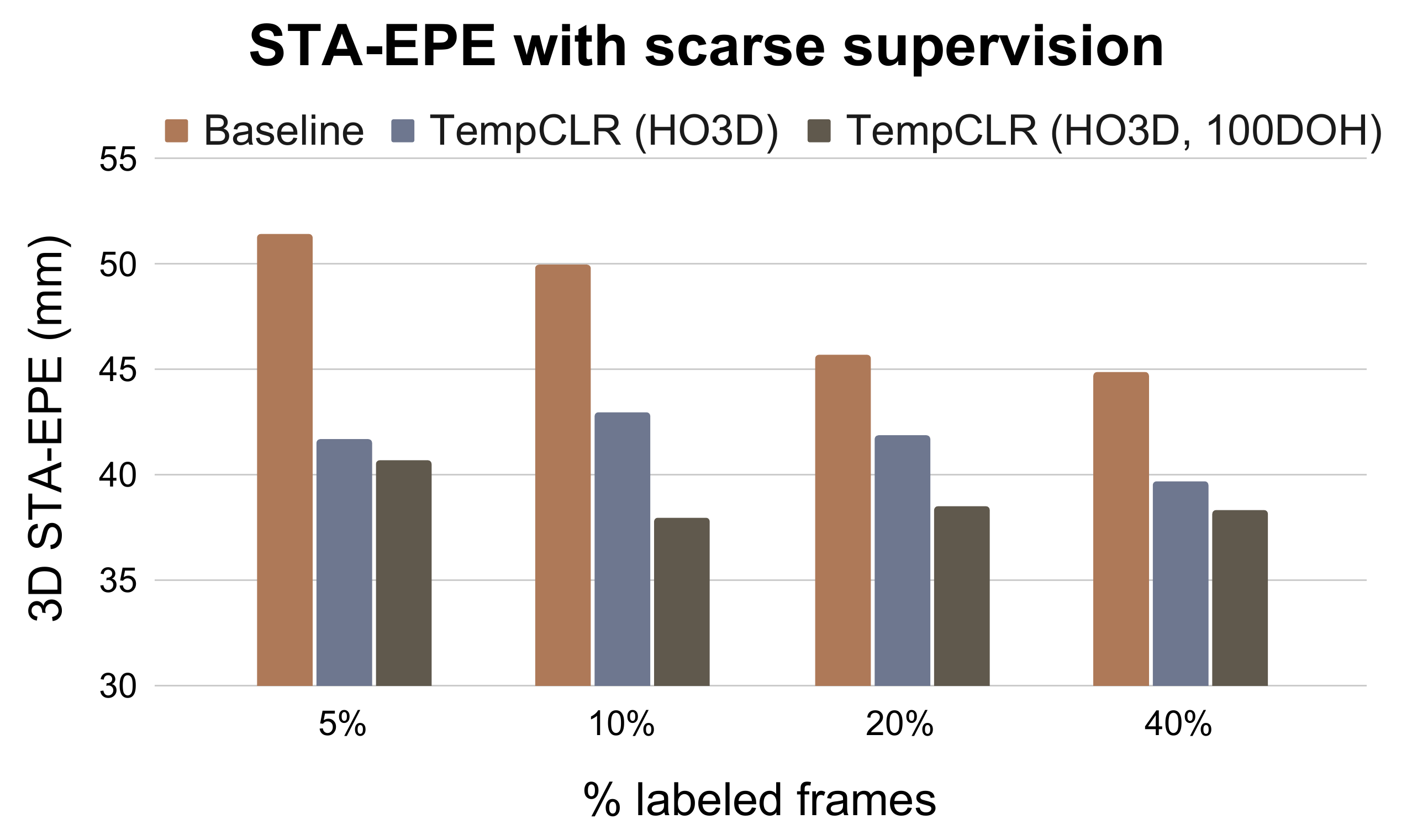}
\end{tabular}   
\vspace{-0.3cm}
\caption{\textbf{Self-supervised performance on \hod}. \methodname achieves better \PAepe (top) and \STAepe (bottom) performances than the fully-supervised baseline ExPose\ccite{choutas2020expose}.
Additional in-the-wild unlabelled data improves \methodname further.}
\label{fig:contrastive_scarce_data}
\vspace{-6mm}
\end{figure}
\begin{figure*}[t]
\centering
\includegraphics[width=1.0\linewidth]{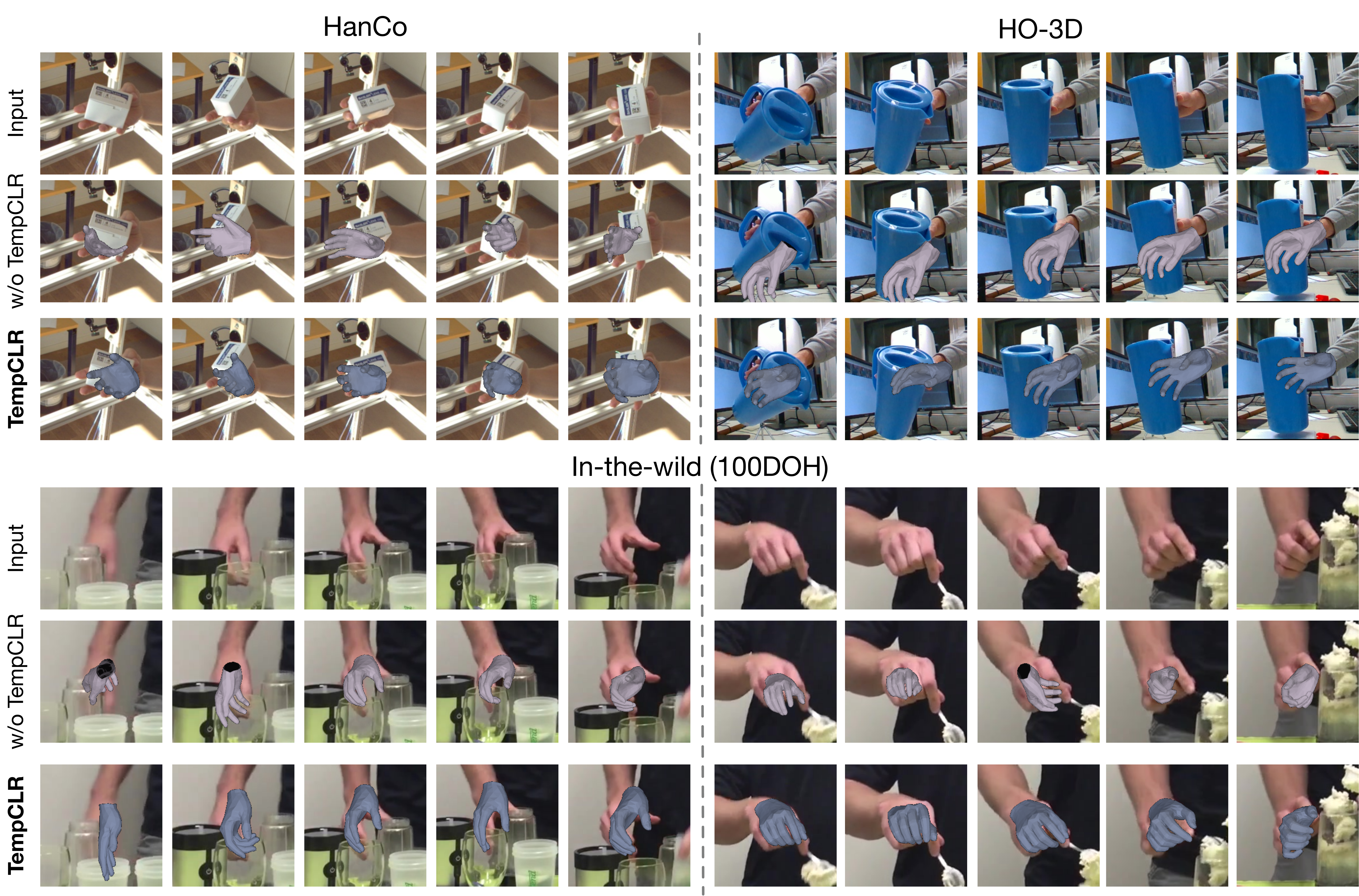}
\vspace{-0.1cm}
\caption{\textbf{Qualitative results} on \hanco\ccite{zimmermann2021contrastive} unlabelled sequences, \hod\ccite{hampali2020honotate} test set, and \inthewild\ccite{shan2020internet} unlabelled sequences. 
Predictions are produced using models described in \refsec{sec:experiments:sota_comparison}. Further qualitative results can be found in \supmat.}
\label{fig:qualitative_res}
\end{figure*}

\myparagraph{Different sampling strategies}
\refTab{tab:sampling_ablation} shows the effects of different sampling strategies. Namely, we compare linear sampling, exponential sampling, and sampling using the absolute value of the hyperbolic tangent function. 
We see that linear sampling leads to the best performance.

\myparagraph{Learning with different amount of supervision}\label{sec:experiments:scarse_data}
We investigate the impact of our pre-training objective with respective to different amount of human-annotated data and \inthewild unlabelled data. 
The ExPose\ccite{choutas2020expose} baseline uses an ImageNet pre-trained encoder.
For our method, we apply \timecontrastive  pre-training either using \hod only, or \hod plus 100DOH to demonstrate the advantage of adding \inthewild data for self-supervised training.
All the hand reconstruction networks are fine-tuned on sparsely annotated sequences from \hod.
We evaluate the performance of the networks on the \hod test set.
\reffig{fig:contrastive_scarce_data} summarizes the results in EPE by progressively increasing the percentage of annotated frames from 5\% to 40\%.
We see that, \methodname consistently improves hand reconstruction by leveraging additional unlabelled data.
Moreover, the use of additional \inthewild unlabelled data (see 100DOH) further improves our performance.
Interestingly, only 20\% of supervised frames are necessary to reach the performance of more densely annotated data. This behaviour is confirmed by\ccite{Hasson2020photometric} and can be explained by the high correlation between neighboring frames of the \hod sequences.

\myparagraph{Window size} When trained on HanCo and fine-tuned on FreiHAND, the PA-EPE error of \methodname with the window sizes $3$, $5$, $15$ are $11.1mm$, $10.9mm$, $11.3mm$, respectively.
Future work could leverage optical flow to detect changes in the sequences for an adaptive window size.

\begin{table}[t]
\resizebox{\columnwidth}{!}{%
\begin{tabular}{ccc}
\toprule
\multirow{2}{*}{\textbf{Method}} & \textbf{HO-3D (train), FH (test)} & \textbf{FH (train), HO-3D (test)} \\
                                 & RA-EPE/PA-EPE                           & STA-EPE/PA-EPE                          \\\hline
Baseline                         & 104.5/18.5                              & 66.1/13.9                               \\
PeCLR                            & 96.0/17.8                               & 62.2/13.6                               \\
TempCLR                          & \textbf{84.6/17.0}                      & \textbf{53.5/13.6}\\
\bottomrule
\end{tabular}}
\caption{\textbf{Cross-dataset evaluation.} 
Methods are trained on HO-3D and evaluated on FH and vice versa.
\methodname generalizes best in both domain shifts. Metrics are in mm.}
\label{tab:cross_dataset_eval}
\end{table}

\subsection{Cross-dataset Evaluation}\label{sec:experiments:cross_eval}
Cross-dataset generalization is rarely reported in the hand reconstruction literature, perhaps because it is widely assumed to be challenging. Yet, it is clearly important for real-world applications.
Given the use of a large amount of unlabelled data for \timecontrastive pre-training, we expect our approach to produce features that are beneficial for generalization on unseen scenes.
To this end, we verify the effectiveness of the models from \refsec{sec:experiments:sota_comparison} in a cross-dataset setting. In particular, we evaluate the performance of the model when trained on FH and evaluated on \hod, and vice versa.
This reveals how the models perform under a domain shift.
\refTab{tab:cross_dataset_eval} reports an improvement over the baseline of $19\%$ in both \RAepe on \freiH and \STAepe on \hod.
These results show that our pre-training objective enables better generalization to unseen scene.
\section{Conclusion}
We introduce, \methodname, a time-contrastive method for hand pose and shape estimation that yields stable \threeD reconstructions through time.
We introduce time-coherent augmentations and probabilistic pair sampling to better account for the temporal information provided by unlabelled videos.
We thoroughly investigate our method, showing that it better captures temporal features and improves reconstruction stability through time. 
We demonstrate that our \methodname achieves \sota results on the \hod and \freiH datasets.
Finally, by means of cross-dataset evaluation, we show the potential of our method's generalization capabilities.

\myparagraph{Acknowledgement} 
Muhammed Kocabas is supported by the Max Planck ETH Center for Learning Systems.
The authors would like to thank Vassilis Choutas for providing the code of the baseline model adopted for the project.

{\small
\balance
\bibliographystyle{ieee_fullname}
\bibliography{config/egbib}
}

%%%%%%%%% TITLE
\title{\methodname: Reconstructing Hands via Time-Coherent Contrastive Learning \textit{*Appendix*}}
\author{Andrea Ziani$^{1*}$ \quad Zicong Fan$^{1,2*}$ \quad Muhammed Kocabas$^{1,2}$ \quad Sammy Christen$^{1}$ \quad Otmar Hilliges$^{1}$\vspace{0.1cm} \\
 $^1$ETH Z{\"u}rich, Switzerland \quad
 $^2$Max Planck Institute for Intelligent Systems, T{\"u}bingen
}

\maketitle
\thispagestyle{empty}

In this document, we first report the implementation details in \refSec{supp:implementation}. 
Next, we provide additional ablation studies and analysis in \refSec{supp:augmentation_abl}.
% We then present a comparison of our \timecontrastive approach with a fully-supervised temporal model in \refSec{supp:exp}. 
Finally, in \refSec{supp:qualitative} we show additional qualitative results and describe, in more details, the failure cases presented in the main text.

\section{Implementation Details}
\label{supp:implementation}
Here we report the details describing the \methodname training procedure, and we clarify how we adjust \peclr\ccite{spurr2021self} to our setting.
We use ResNet\ccite{he2016resnet} as the backbone, which takes monocular \rgb images of size $224\times224$ as input. 
We employ Adam\ccite{kingma2015adam} optimizer for training. 

\myparagraph{Time-contrastive pre-training}
For this pre-training stage, 
we train the model with batches of size $2048$ and a learning rate of $4.5e^{-3}$.
A linear warmup is performed for the first 10 epochs. 
After that, we use cosine annealing for the remaining training iterations. 
We train for a total of $50$ epochs, which we found to  perform the best empirically. 
For pre-training with multiple datasets, we perform a sampling strategy to balance the samples, such that there is an equal amount of samples from each dataset.

To select positive frames for the contrastive training, the window size is set to $15$ for \hod and \doh.
For \hanco\ccite{zimmermann2021contrastive}, because the frame rate is not available and the number of frames per sequence is much smaller compared to other datasets, we fix the positives' window size to $5$.
For \methodname geometric agumentation, we augment the sampled images using rotation $r \in [-45^\circ, 45^\circ]$, scaling $s \in [0.6, 2.0]$, and translation $t \in [-\text{im\_size} \times 0.3, \text{im\_size}$ $\times 0.3]$ in pixel. 
The randomness of these augmentations is fixed per sequence, such that the same transformation is applied to each frame.
In addition, we apply random appearance transformation, independently to each frame, in terms of channel noise $n \in [0.6, 1.4]$, sobel filter with a kernel size of 3, and color drop.

\myparagraph{Supervised fine-tuning}
The fine-tuning is performed until convergence based on validation performance, which we leave out before training on the supervised datasets.
We fine-tune our model with a learning rate of $5e^{-4}$ in conjunction with a cosine annealing scheduler. The batch size is set to $128$. 
In this stage, we augment the data using only geometric transformations. In particular, we use rotation $r \in [-90^\circ, 90^\circ]$, scaling $s \in [0.7, 1.3]$, and translation $t \in [-\text{im\_size} \times 0.4, \text{im\_size} \times 0.4]$ in pixel.

\myparagraph{PeCLR adaptation}
To fairly compare \methodname with \peclr\ccite{spurr2021self}, we adapt its contrastive training to our model-based hand pose estimation architecture.
In particular, in the pre-training stage, we use the same contrastive training described in \peclr, where geometric transformations applied on the images are reversed in the latent space to achieve the equivariance property. 
Then, in the fine-tuning stage, we add the hand reconstruction architecture (Sec. 3.2 of the main text) in place of the model-free decoder originally used by Spurr \etal \ccite{spurr2021self}.

\begin{figure}[t]
\centering
\includegraphics[width=\linewidth]{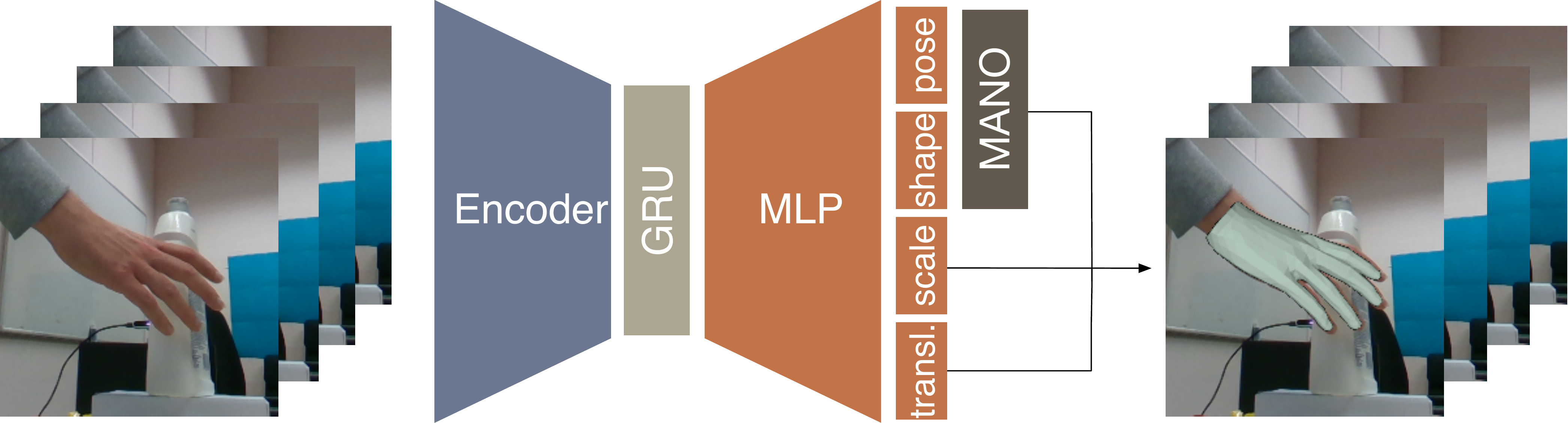}
\caption{\textbf{Temporal model architecture. }In contrast to the architecture employed by \methodname, an additional recurrent layer is added between the encoder and the decoder.}
\label{fig:temporal_model}
\end{figure}
\section{Experiments and Analysis}
\label{supp:augmentation_abl}
Here, we report additional experiments and analysis.
Then we compare \methodname with a temporal model.

\begin{figure}[t]
    \centering
    \includegraphics[width=1.0\linewidth]{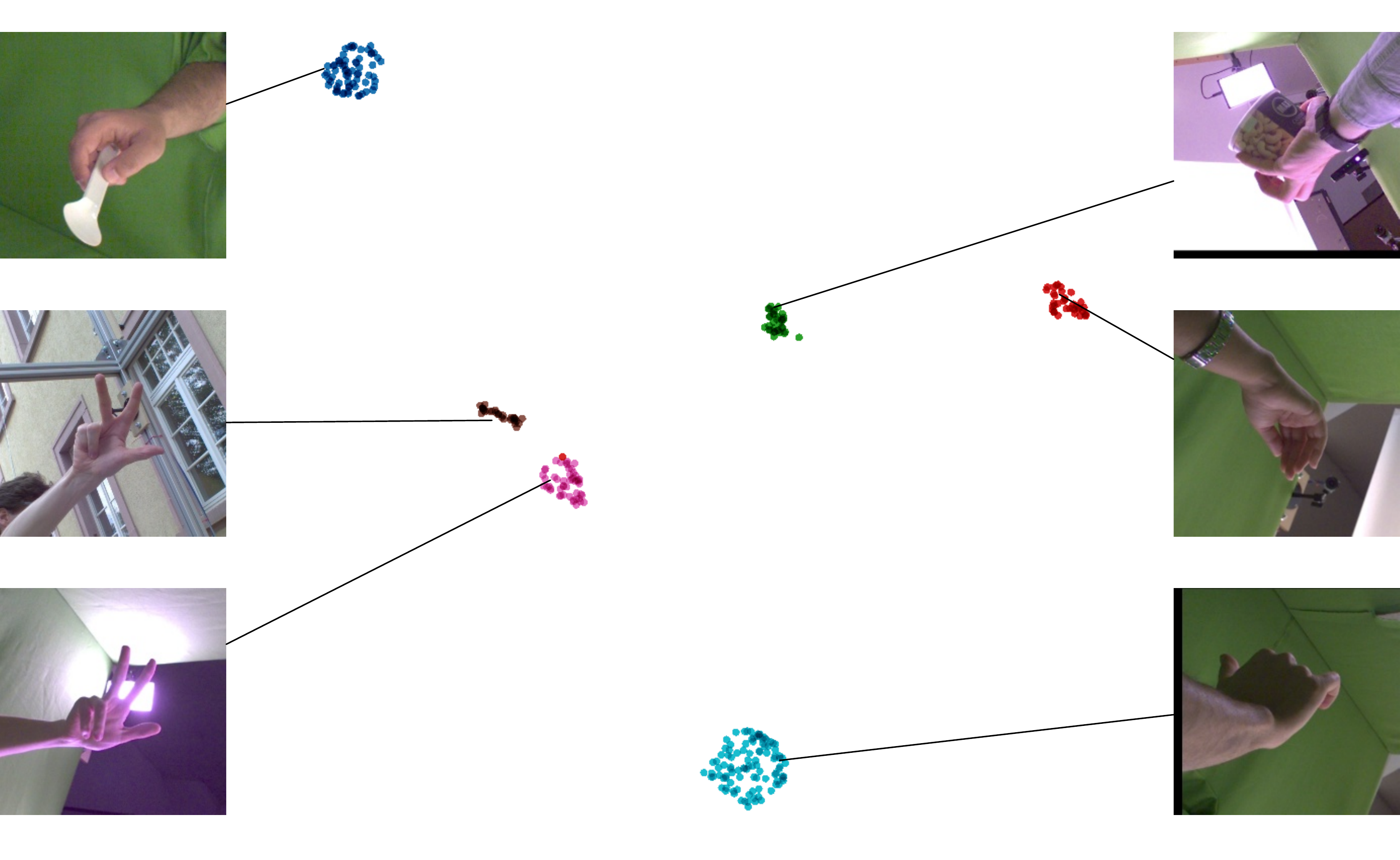}
    \caption{t-SNE embeddings from a \methodname pre-trained encoder. Seqs. with similar actions (bottom-left) are closer.}
    \label{fig:tsne_similar_pose}
\end{figure}

\myparagraph{Negative samples with similar actions}
Our contrastive formulation aims at learning an embedding space such that similar hand poses are closer in the space, which includes the orientation and global translation of the hands.
For example, although the actions of ``picking up" and ``placing back" a cup are often similar in hand poses, it is unlikely for them to have exactly the same orientation and translation. 
To provide qualitative evidence, we performed a t-SNE projection of six different HanCo sequences, two of them of a similar action. \reffig{fig:tsne_similar_pose} shows that the embeddings of videos performing similar actions  are closer in the projection space but do not overlap.
One advantage of \methodname is to leverage large-scale unlabelled video data with very diverse and variable hand poses as opposed to quasi-static grasping actions. 
In this more realistic setting, it is less likely for the negative samples to have exactly the same hand pose, global orientation, and translation as the positive samples.

\myparagraph{Comparison with a temporal model}
We compare our single-frame \methodname model trained on HO-3D with time-contrastive pre-training, against a temporal model similar to VIBE\ccite{kocabas2020vibe}.
Since there is no large-scale archive of hand motion captures to train the discriminator part of the original VIBE architecture, we present the results without the motion discriminator of VIBE.
Note that we already provided a comparison with existing temporal models in Tab. 1 and Tab. 2 of the main paper.

\refFig{fig:temporal_model} demonstrates the temporal model.
A sequence of frames $I_1,\cdots , I_t$ is fed into a ResNet18\ccite{he2016resnet} encoder, which functions as a feature extractor and outputs a vector $f_i$ for each frame. These feature vectors $f(I_1), ..., f(I_t)$ are sent to a Gated Recurrent Unit\ccite{cho-etal-2014-learning} (GRU) layer which yields a latent feature vector $g_i$ for each frame, $g(f(I_1)), ..., g(f(I_t))$, based on the
previous frames. Then, each of these latent vectors are fed into $T$ regressors with iterative feedback as in\ccite{choutas2020expose}. 
The training loss function is a linear combination of 2D re-projection loss, 3D joint errors, and pose and shape parameter loss:
\begin{gather}
L = \lambda_{2\text{D}} L_{2\text{D}} + \lambda_{3\text{D}} L_{3\text{D}} + \lambda_{\Theta} L_{\Theta},\\
L_{2\text{D}} = \norm{J^{2\text{D}} - \hat{J}^{2\text{D}}}_1 ,  L_{3\text{D}} = \norm{J^{3\text{D}} - \hat{J}^{3\text{D}}}_1 ,\\  L_{\Theta} = \norm{\{\theta, \beta\} - \{\hat{\theta}, \hat{\beta}\}}^{2}_2.
\end{gather}
All the variables with a hat denote the network predictions, while all the variables without a hat denote the ground truth. Moreover, $J^{2\text{D}} \in \mathbb{R}^{21 \times 2}$ and $J^{3\text{D}} \in \mathbb{R}^{21 \times 3}$ represent the twenty one 2D and 3D keypoints, respectively.

We compare our \timecontrastive approach with the fully-supervised ExPose\ccite{choutas2020expose} baseline and the temporal model, solely trained on \hod labelled sequences. 
\refTab{tab:temporal_model} shows that the temporal model reaches better performances compared to the single-frame architecture. However, our \timecontrastive approach, without employing any additional in-the-wild data during pre-training, improves the performance over the temporal architecture by $4.5\%$ in \PAvtov and \PAepe.
\begin{table}[t]
\centering
\resizebox{\columnwidth}{!}{
\begin{tabular}{lccccc} 
 \hline
 \multirow{2}{*}{Method} & PA-V2V  & PA-EPE & F@5 & F@15\\
 & ($mm$) $\downarrow$ & ($mm$) $\downarrow$ & $mm$ $\uparrow$ & $mm$ $\uparrow$\\
 \hline\hline
 Baseline\ccite{choutas2020expose} & 12.6 & 12.7 &  0.389 & 0.905 \\
 \rowcolor{gray!15} Temporal Model & 11.1 & 11.2 &  0.447 & 0.929 \\
 \textbf{TempCLR (ours)} & \textbf{10.6} & \textbf{10.6} & \textbf{0.481} & \textbf{0.937}\\
 \hline
\end{tabular}
}
\caption{\textbf{\methodname comparison with the temporal model on \hod\ccite{hampali2020honotate}.} Our method outperforms both single-frame baseline and the temporal model architecture on all reported metrics. This indicates that our \timecontrastive pre-training helps reconstructing more accurate hands in a heavy occlusion scenario.
}
\label{tab:temporal_model}
\end{table}

\myparagraph{Comparison with different backbones.}
\refTab{tab:deeper_backbones} shows the performance of \methodname with different backbones. In particular, to analyse whether our method can be applied on deep architectures we report results using ResNet18, ResNet50, ResNet101, and HRNet w48 backbones.
The experiment shows that \methodname performs better consistently employing deeper backbones.
However, since most existing methods use ResNet18 and ResNet50, we use those backbones for \methodname in the main paper for fair comparison.

\begin{table}[t]
\centering
\resizebox{\columnwidth}{!}{
\begin{tabular}{lccccc} 
 \hline
 \multirow{2}{*}{Method} & PA-V2V  & F@5 & F@15\\
 & ($mm$) $\downarrow$ & $mm$ $\uparrow$ & $mm$ $\uparrow$\\
 \hline\hline
 \multicolumn{4}{c}{HO-3D}\\
 \hline\hline
  TempCLR-18& 10.0 &  0.505 & 0.947\\
  TempCLR-101& 10.0 &  0.507 & 0.947\\
  TempCLR-HRNet& 
  10.0 & 0.512 & 0.943\\
 \hline\hline
 \multicolumn{4}{c}{FreiHAND}\\
 \hline\hline
 TempCLR-18 &
 10.9  & 0.513 & 0.930 \\
 TempCLR-50 &  10.2 & 0.541 & 0.941 \\
 TempCLR-101 &  10.0 & 0.543 & 0.944 \\
 \hline
\end{tabular}
}
\vspace{-0.2cm}
\caption{\textbf{Comparison of TempCLR on various backbones on HO-3D and FreiHAND.}}
\label{tab:deeper_backbones}
\end{table}
\begin{figure*}[t]
\centering
\includegraphics[width=0.60\linewidth]{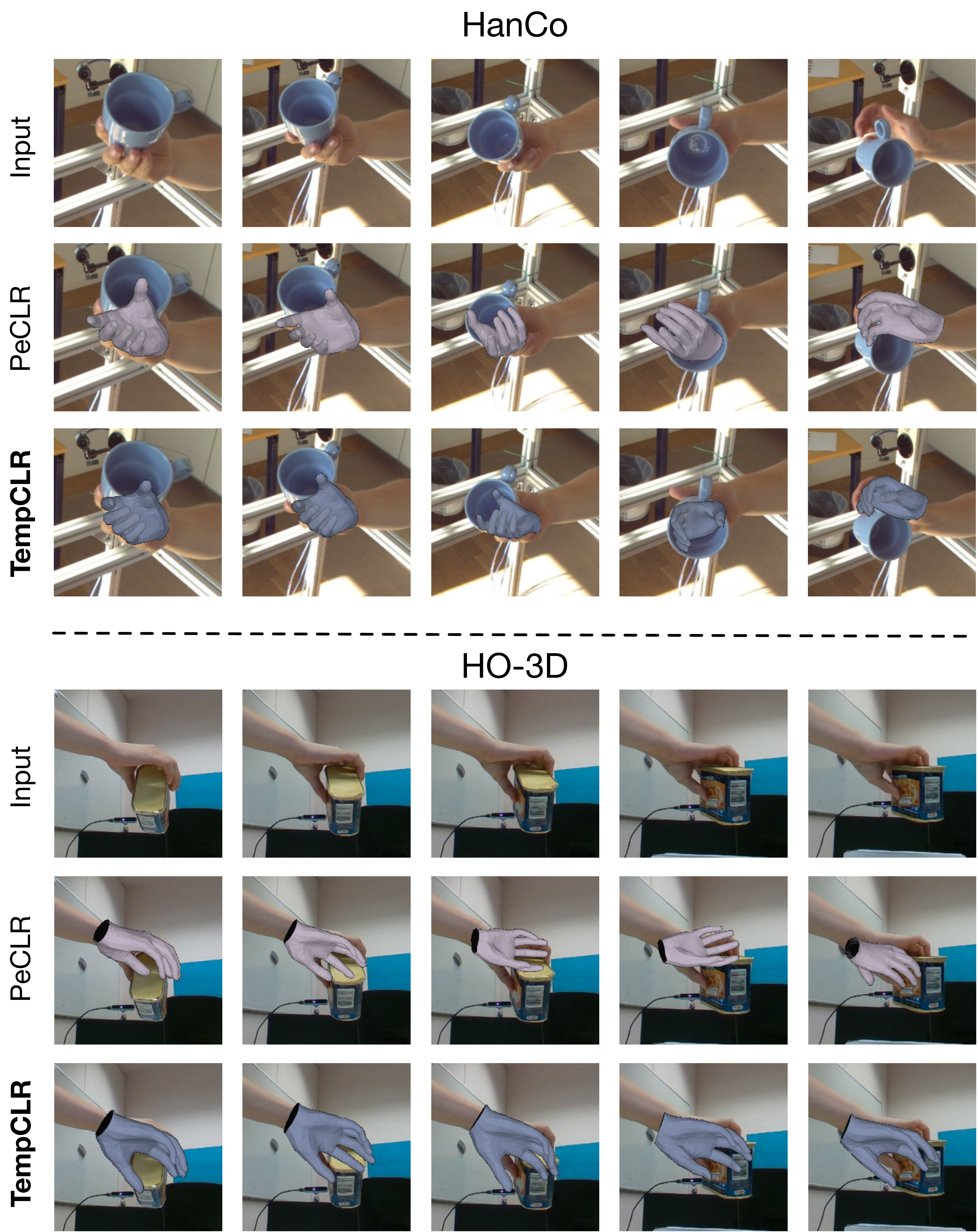}
\vspace{-0.2cm}
\caption{\textbf{\methodname compared with \peclr \ccite{spurr2021self}.} Predictions are shown on unlabelled sequences from \hanco\ccite{zimmermann2021contrastive} and on the \hod\ccite{hampali2020honotate} test set. TempCLR reconstructions are smoother over time when compared to \peclr in a heavy occlusion scenario. }
\label{fig:peclr_comparison}
\end{figure*}

\section{Qualitative Results and Failure Cases}
\label{supp:qualitative}
In this section we compare the qualitative results from \methodname and \peclr\ccite{spurr2021self}. 
Next, we analyze the limitations of \methodname.
Finally, we provide additional qualitative results (\reffig{fig:supp_qualitative_res_hanco}-\ref{fig:supp_qualitative_res_itw}) produced by our baseline model, PeCLR, and \methodname.

\begin{figure*}[t]
\centering
\includegraphics[width=0.80\linewidth]{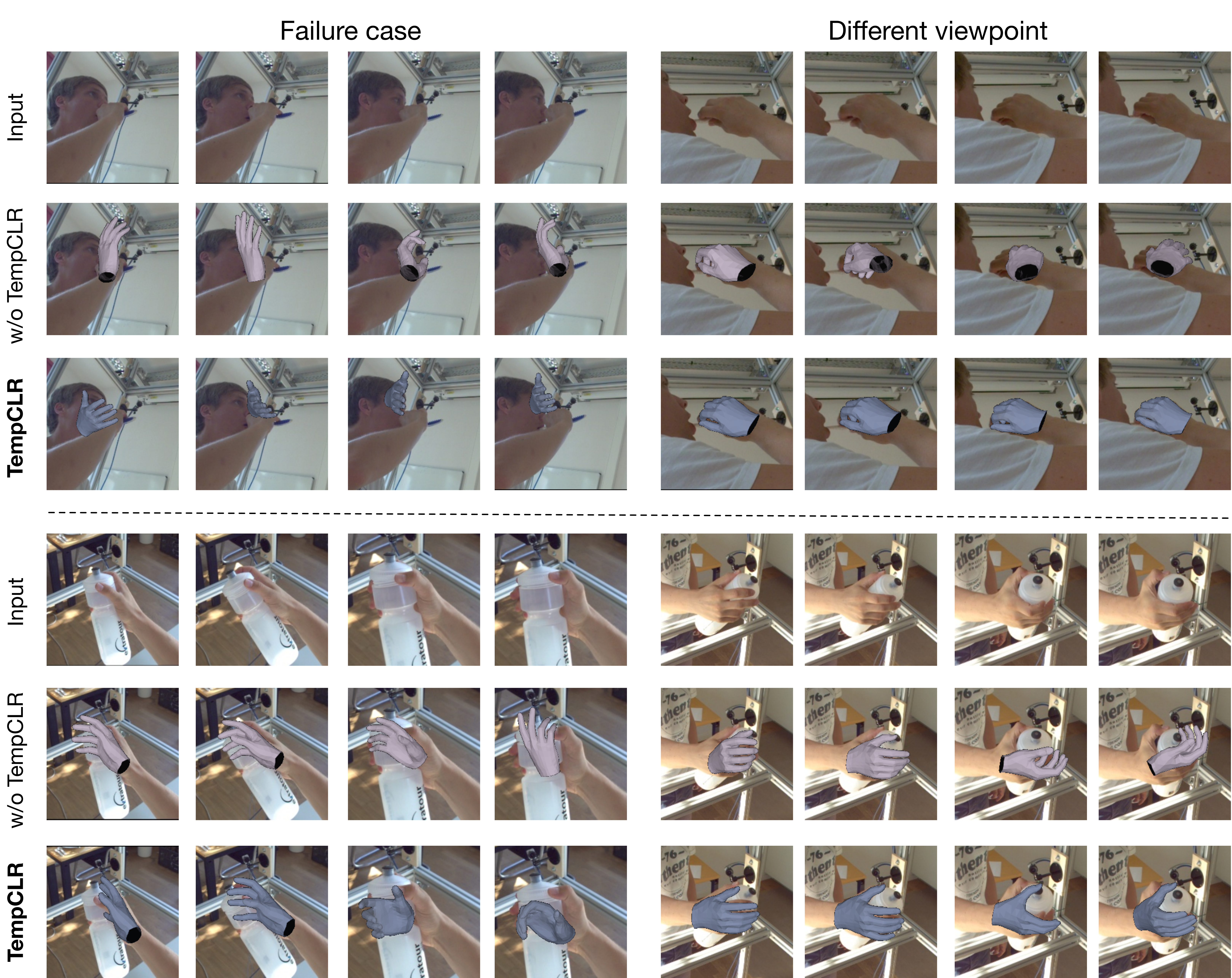}

\vspace{-0.1cm}
\caption{\textbf{\methodname failure cases.} In the sequence on top we observe incorrect reconstructions when the hand is not clearly visible for almost an entire sequence. In the sequence at the bottom, an uncontrollable variation in image scale between second and third frame leads to incorrect hand pose predictions. The fully-supervised baseline\ccite{choutas2020expose} also struggles with these sequences.}
\label{fig:failure_cases}
\end{figure*}

\myparagraph{Compare \methodname and \peclr}
Section 4 of the main text shows an improvement of \methodname over \peclr by $7.4\%$ in \PAvtov on \hod\ccite{hampali2020honotate}.
To build an intuition for such an improvement, we show qualitative results from \methodname and \peclr.
\refFig{fig:peclr_comparison} shows results over sequences from both \hod\ccite{hampali2020honotate} and \hanco\ccite{zimmermann2021contrastive}, obtained as described in Section 4.4 of the main paper. 
We observe that both approaches predict similar hand poses and shapes. However, results from \methodname are smoother and more coherent in time than \peclr, which is expected due to our \timecontrastive training.
To quantify this, we measure the acceleration error in the single-frame reconstruction. The acceleration error of \methodname, PeCLR, and the baseline are $36.94$, $38.51$, $52.36$ (in $mm/{s^2}$) respectively when trained on FH and evaluated on HO-3D training set with a ResNet50 backbone.

In particular, the goal of \peclr contrastive training is to attract the embeddings of different transformations of an image representing the same hand pose. However, artificial variations (\eg rotation, scaling, translation) are not expressive enough to account for all the possible changes within consecutive frames. 
All of these small differences in the image space may actually lead to significantly different latent representations.
For example, consider two consecutive frames of a sequence, where the hand is more occluded by an object in the second frame.
Since there is no temporal component in \peclr, these slightly different but similar hand poses are likely to be repelled during training.
On the other hand, the \methodname contrastive objective accounts for this small variations of the images by attracting the latent representation of frames closer in time.
Hence, as shown in Fig. 4 of the main text, images with similar hand poses clustered together in the \methodname latent space, leading to coherent hand reconstructions.

\myparagraph{\methodname failure cases}
\refFig{fig:failure_cases} shows the failure cases of our model.
First, \methodname fails when the hand is heavily occluded (see the top split of \reffig{fig:failure_cases}).
Given a sequence of images with not clearly visible hand poses, there could be multiple potential hand poses for the occluded region. 
Since the underlying hand poses are ill-defined, these images might make the pre-training objective less defined.

Further, \methodname fails when there is a drastic variation in image scale across a video (see the bottom split of \reffig{fig:failure_cases}).
This problem arises since our time-coherent augmentation strategy does not account for these unexpected geometric transformations happening over time.
A solution could be to separate a sequence into sub-sequences by detecting sudden changes in images. For example, optical flow could be used to measure such changes. 
In this way, the attraction of latent representations of frames with different geometric variations would be avoided and the motion cues of the video preserved.

\begin{figure*}[t]
\centering
\includegraphics[width=.70\linewidth]{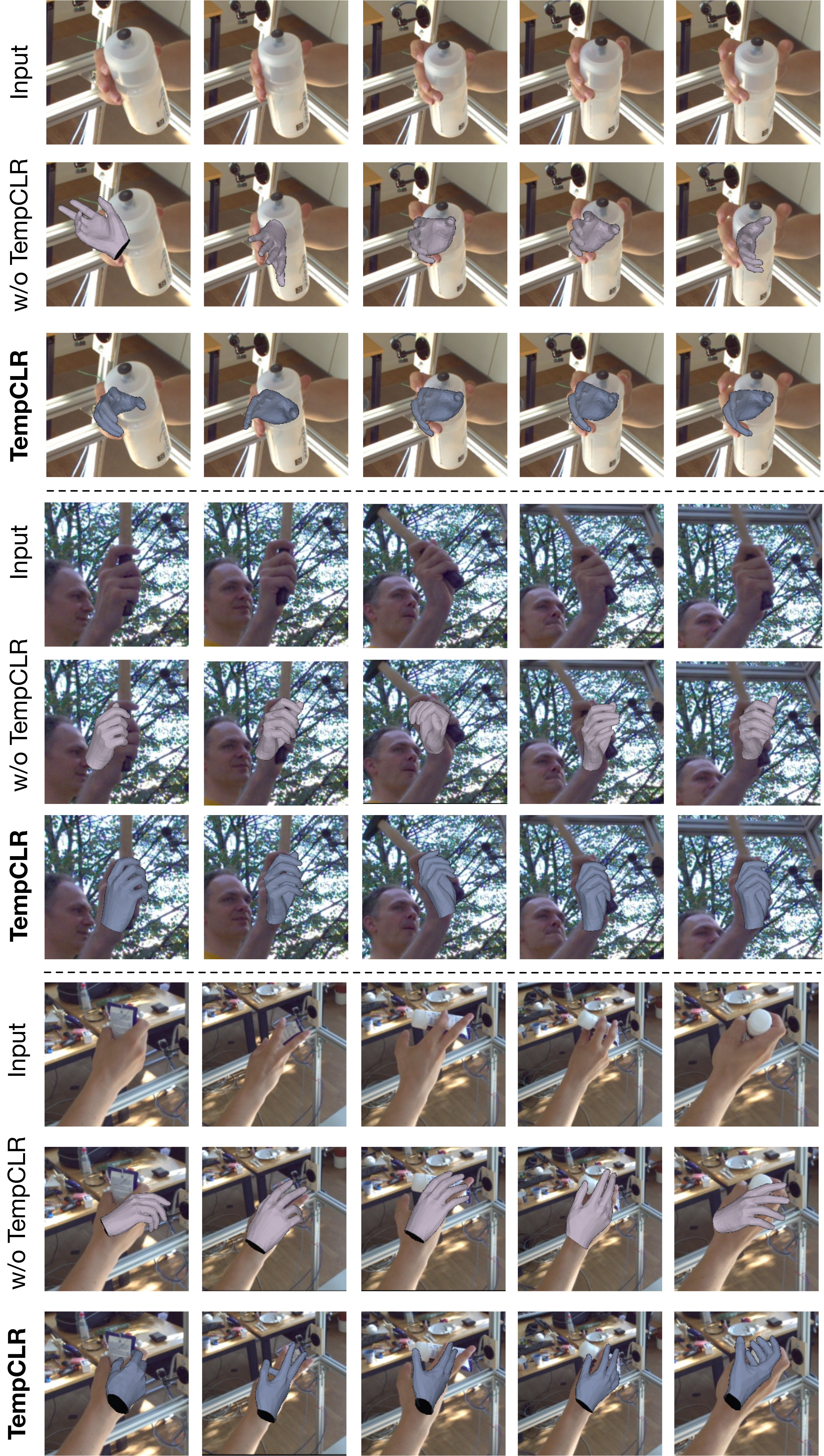}

\caption{\textbf{Qualitative results on the \hanco\ccite{zimmermann2021contrastive} dataset.} Predictions are shown for the fully-supervised baseline \cite{choutas2020expose} and \methodname. For \methodname, pre-training is  carried out on unlabelled \hanco sequences. Both models are fine-tuned on \freiH\ccite{zimmermann2019freihand}.
Note that the ground-truth is not publicly available, hence we only visualize the predictions.}
\label{fig:supp_qualitative_res_hanco}
\end{figure*}

\begin{figure*}[t]
\centering
\includegraphics[width=.70\linewidth]{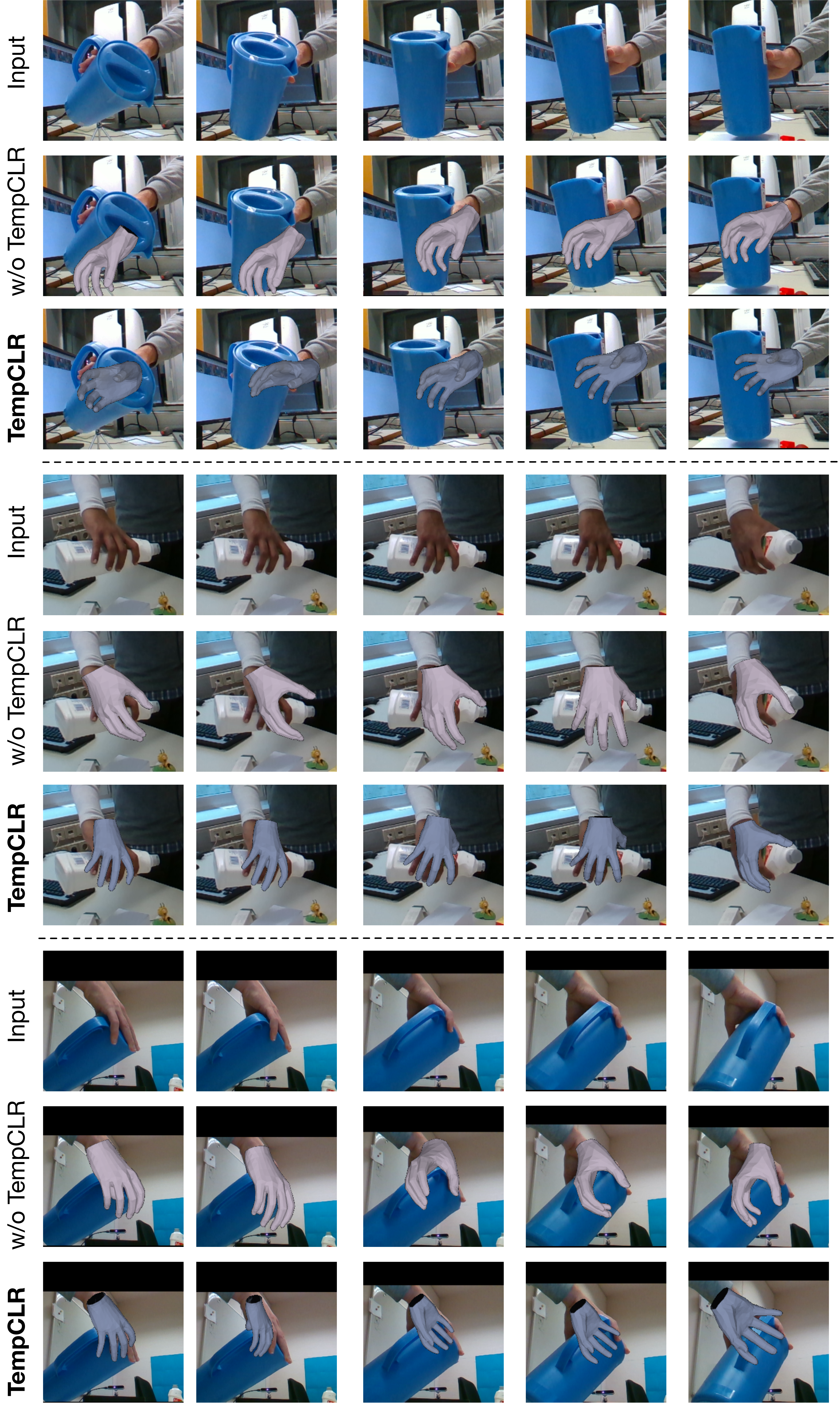}

\caption{\textbf{Qualitative results on the \hod\ccite{hampali2020honotate} test set.} Predictions are shown for the fully-supervised baseline \cite{choutas2020expose} and \methodname. For \methodname, pre-training is  carried out on unlabelled \hod sequences. Both models are fine-tuned on \hod\ccite{hampali2020honotate}.
Note that the ground-truth of the test set is not publicly available, hence we only visualize the predictions.}
\label{fig:supp_qualitative_res_ho3d}
\end{figure*}

\begin{figure*}[t]
\centering
\includegraphics[width=.70\linewidth]{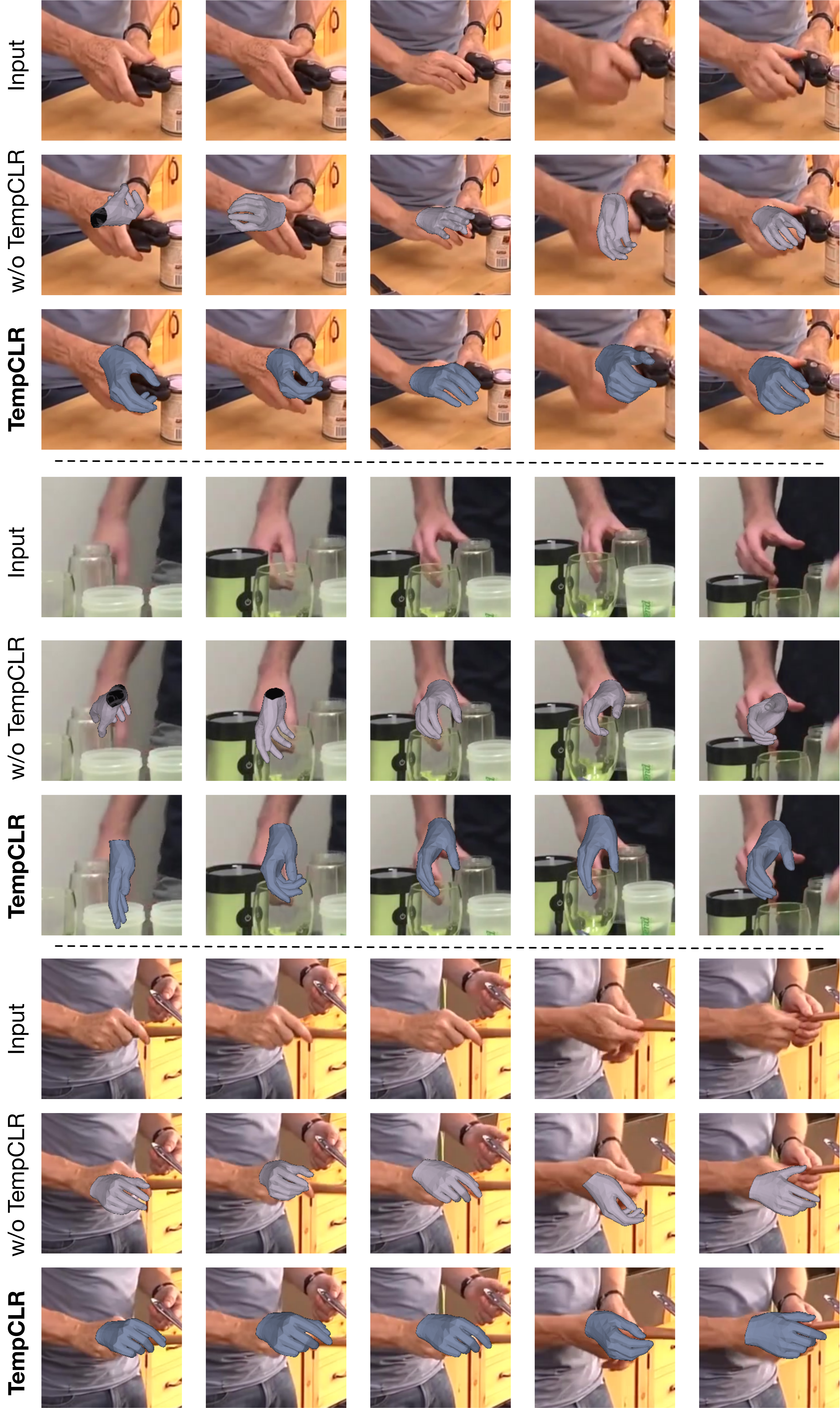}

\caption{\textbf{Qualitative results on \inthewild (\doh\ccite{shan2020internet}) sequences.} Predictions are shown for the fully-supervised baseline \cite{choutas2020expose} and \methodname. For \methodname, pre-training is carried out on unlabelled \hanco sequences. Both models are fine-tuned on \freiH\ccite{zimmermann2019freihand}.
Note that the ground-truth is not publicly available, hence we only visualize the predictions.}
\label{fig:supp_qualitative_res_itw}
\end{figure*}

\end{document}